%% file: conf-text.tex
\documentclass[sigconf,screen]{acmart} 

\usepackage{listings}

\setcopyright{acmcopyright}
\acmPrice{15.00}
\acmDOI{10.1145/3136755.3143022}
\acmYear{2017}
\copyrightyear{2017}
\acmISBN{978-1-4503-5543-8/17/11}
\acmConference[ICMI'17]{19th ACM International Conference on Multimodal Interaction}{November 13--17, 2017}{Glasgow, UK}

\begin{document}
\title{MIRIAM: A Multimodal Chat-Based Interface for Autonomous Systems}

\author{Helen Hastie}
\author{Francisco Javier Chiyah Garcia}
\author{David A. Robb}
\affiliation{
 \institution{Heriot-Watt University}
  \city{Edinburgh} 
  \country{UK} 
  \postcode{EH14 4AS}
}
\email{h.hastie, fjc3, d.a.robb@hw.ac.uk}    

\author{Pedro Patron}
\author{Atanas Laskov}
\affiliation{
  \institution{SeeByte Ltd}
  \city{Edinburgh} 
  \country{UK} 
  \postcode{EH4 2HS}
}
\email{pedro.patron, atanas.laskov@seebyte.com}
\renewcommand{\shortauthors}{H.Hastie, F.J.C.Garcia, D.A.Robb, P.Patron and A.Laskov}


\begin{abstract}
We present MIRIAM (Multimodal Intelligent inteRactIon for Autonomous systeMs), 
a multimodal interface to support situation awareness of autonomous vehicles through chat-based interaction. The user is able to chat
about the vehicle's plan, objectives, previous activities and mission progress. The system is mixed initiative in that it pro-actively sends messages about key events, such as fault warnings. We will demonstrate MIRIAM using SeeByte's SeeTrack command and control interface and Neptune autonomy simulator.

 \end{abstract}




\maketitle

\input{body-conf}

\begin{acks}
This research was funded by DSTL (MIRIAM/ACC101939); and the RAEng/Leverhulme Trust Senior Research Fellowship Scheme (Hastie/LTSRF1617/13/37). We acknowledge Prof. Y Petillot, Dr X Liu and Dr Z Wang for their input and guidance.
\end{acks}

\bibliographystyle{ACM-Reference-Format}
\bibliography{body-conf} 

\end{document}

%% file: body-conf.tex
\section{Introduction}
There has been a recent upsurge in automated conversational agents or chatbots, performing a variety of functions such as resolving users' technical problems \cite{serban}; performing tasks in the real world (e.g. ordering flowers); and more recently pure social interaction \cite{vinyals2015neural}. These systems run on company websites e.g. for customer services, or on third party instant messaging (IM) apps (e.g. Facebook/Skype/Whatsapp). Whilst multimodality has been used to augment  modes of input and output, there has been little prior work on multimodal interfaces, whereby the user can discuss with the system remote, on-going activities and events shown through multimedia output, thus contextualising interaction and increasing the
conversational bandwidth of the conversational agent.   

We present MIRIAM (Multimodal Intelligent inteRactIon for Autonomous systeMs): a prototype multimodal interface that combines an automated chatbot system with a command and control (C2) interface. This interface allows for observation and control of 
unmanned systems (UXVs).
The vehicles are controlled by SeeByte's Neptune autonomy framework, a vehicle-agnostic platform supporting 
multi-vehicle missions from different domains (air, surface, water).
Neptune autonomy framework can adapt the mission plan dynamically during execution
based on sensed environmental information, such as the water depth, obstacles or detected objects of interest.

Human-human chat through instant messaging has been used in a variety of C2 situations to improve situation awareness, particularly in time-critical situations \cite{medinaetal2007}. However, there has been little use of automated chatbots in this domain. A multimodal interface such as MIRIAM is key to successful deployment of autonomous systems, as it increases transparency of autonomous actions and provides tactical-situation-assessment, resulting in fewer unnecessary mission aborts. Refer to the project website for more details and to request a login \url{http://www2.macs.hw.ac.uk/miriam/interface}.


\section{Demonstration}

\begin{figure}[t]
\centering
\includegraphics[width=0.9\linewidth]{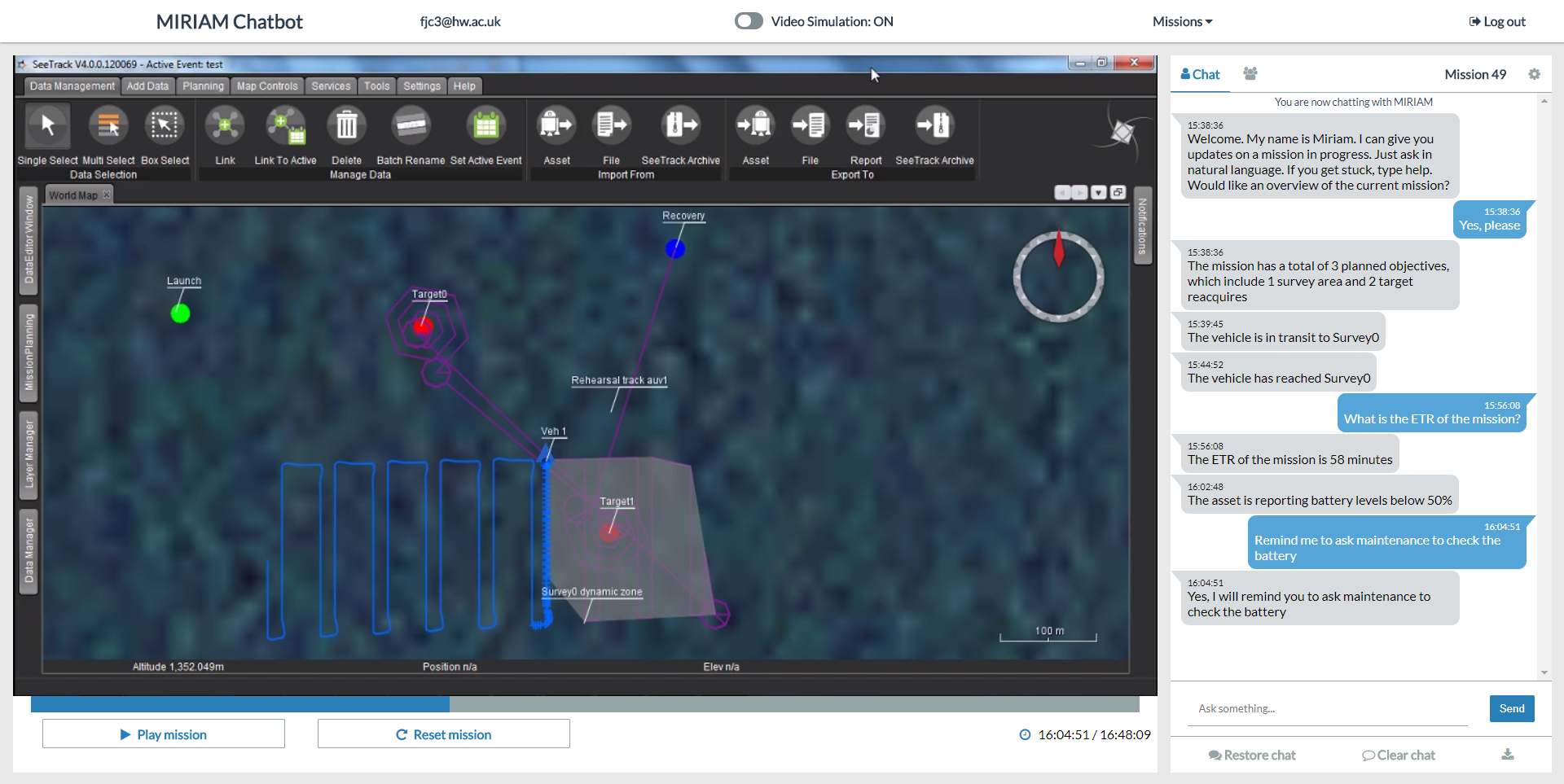}
\caption{Multimodal chat-based interface for UXVs \copyright Heriot-Watt University \copyright SeeByte} \label{fig:miriamchatbot}
\end{figure}

The MIRIAM interface will demonstrate a mission whereby the conference delegates will be able to interact and chat about an underwater autonomous vehicle's activity, as illustrated in Figure \ref{fig:miriamchatbot}. The multimodal interface shows the vehicle track and progress on the left-hand-side through SeeByte's SeeTrack interface and includes a chat-interface of the right-hand-side. Possible natural language interactions include asking about the vehicle's current status, the plan and its current objectives, estimated time of arrival at specific locations, previous activities, mission progress, hardware fault diagnosis, and estimated time of completion of a specified objective. The system is mixed initiative in that it pro-actively sends warnings about important events, such as vehicle faults, critical battery status or a change in objectives. MIRIAM recognises different levels of importance and will pin critical alerts. The interface can be customised for various user-preferences and allows for the creation of reminders through chat. 

\section{System Architecture}
\begin{figure}[t]
\centering
\includegraphics[width=0.7\linewidth]{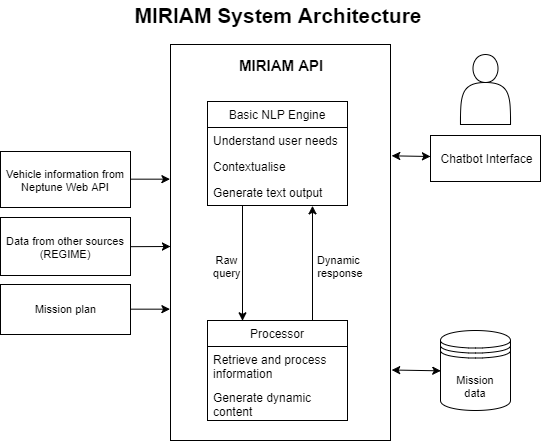}
\caption{System architecture} 
\label{fig:miriamsys}
\end{figure}

The system architecture is illustrated in Figure \ref{fig:miriamsys}. Three sources of data are used as input: 1) the mission plan; 2) vehicle status as outputed through an API from SeeByte's vehicle Neptune software; and 3) mission reports as generated by a system called REGIME described in \cite{hastieicmi2016demo,hastieicmi2017trust}. MIRIAM parses, organises and stores this information into an SQL database for use by the conversational agent.

Internally, the system has two main components. Firstly, an NLP Engine contextualises and parses the user's input for intent, formalising it as a semantic representation. This component is a combination of AIML (Artificial Intelligence Mark-up Language) and a custom-built parser that can process both static data based on keywords and pattern matching, as well as dynamic data, such as names and mission-specific words. For example, the chatbot can recognise ``auv1'', the particular name given to a vehicle in the mission plan, without the requirement to hard-code this name into the dictionary of keywords. 
This component can handle anaphoric references over multiple utterances  e.g. ``Where is Survey0?'' ... ``What time did it finish?''. It also handles ellipsis e.g.``Where is Survey0?'' ....``What about Survey1?''. Secondly, the Processor Component shown in Figure \ref{fig:miriamsys} uses the semantic representation of the user's input to create a suitable reply by retrieving the relevant information from the database.

\begin{figure}[t]
\centering
\includegraphics[height=0.35\textheight]{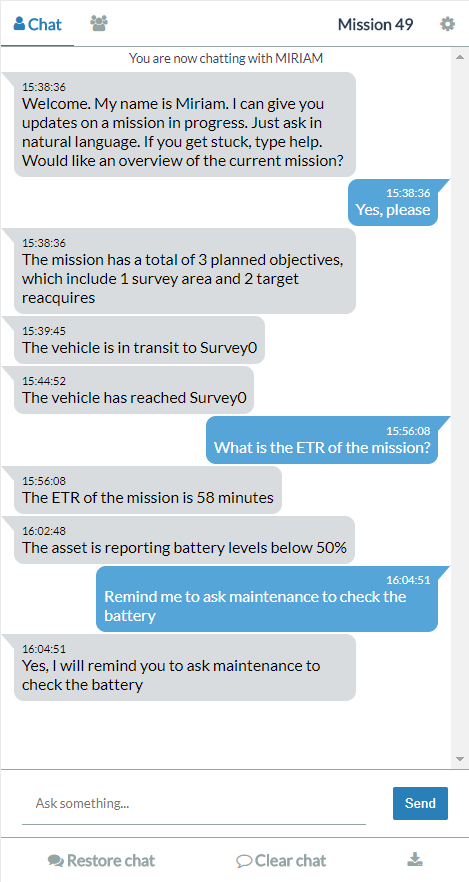}
\caption{Example dialogue from Figure \ref{fig:miriamchatbot}} 
\label{fig:miriamchatzoom}
\end{figure}

As illustrated in  Figure \ref{fig:miriamchatbot}, the operator is able to track the mission progress and assets visually, whilst chatting (see Figure \ref{fig:miriamchatzoom} for a zoomed view of the example dialogue). The prototype system we will present uses a video of a simulated mission to demonstrate the chat functionality without the need to connect to a real UXV.  

In a future version, a direct communication link to the SeeTrack Neptune autonomy monitoring interface will be implemented, allowing for real-time data flow between the chatbot and the UXVs. This will enable a wider range of contextual information through multimodal input (e.g. clicking on the vehicle in question) and through multimodal output (e.g. highlighting the survey area in discussion).  Other contextual information could also be taken into account such as the cognitive load of the user for intelligent alerting \cite{cummings04}. 
Future work  also entails making the system able to discuss multiple vehicles and explain causal reasoning behind its autonomous actions. 


